%% file: main.tex
\documentclass[10pt,twocolumn,letterpaper]{article}

\usepackage{iccv}              
\usepackage{multirow}


\definecolor{iccvblue}{rgb}{0.21,0.49,0.74}
\usepackage[pagebackref,breaklinks,colorlinks,allcolors=iccvblue]{hyperref}

\def\paperID{5684} 
\def\confName{ICCV}
\def\confYear{2025}

\title{RASA: Replace Anyone, Say Anything – A Training-Free Framework for Audio-Driven and Universal Portrait Video Editing}

\author{Tianrui Pan$^{1,2}$ \quad Lin Liu$^{2,\dagger}$ \quad Jie Liu$^{1,\dagger}$ \quad Xiaopeng Zhang$^{2}$, \\  
Jie Tang$^{1}$ \quad Gangshan Wu$^{1}$ \quad Qi Tian$^{2}$ \\[0.3cm] $^1$State Key Laboratory for Novel Software Technology, Nanjing University\\ $^2$Huawei Inc.
}

\begin{document}
\maketitle

{
\let\thefootnote\relax\footnotetext{\noindent$^\dagger$Corresponding author.}
}

\input{sec/0_abstract}    
\input{sec/1_intro}

\begin{figure*}
    \centering
    \includegraphics[width=1.0\linewidth]{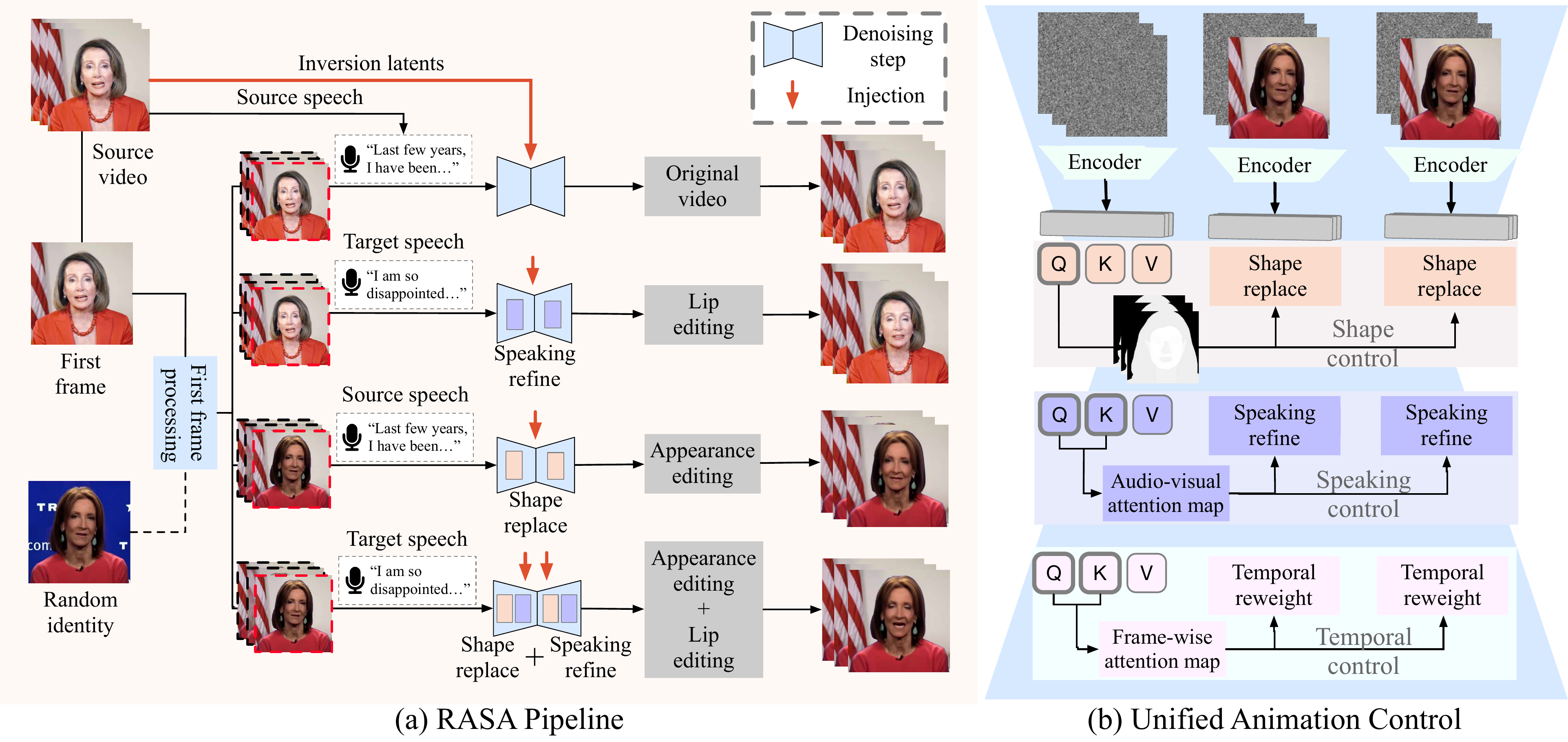}
    \caption{\textbf{The overall structure of RASA.} The left part (a) illustrates the tuning-free universal portrait video editing pipeline. The right part (b) details the denoising steps with unified animation control, outlining the process for achieving multi-task portrait video editing.}
    \label{fig:model_structure}
\end{figure*}

\input{sec/2_relatedwork}
\input{sec/3_method}
\input{sec/4_experiments}
\input{sec/5_conclusion}
{
    \small
    \bibliographystyle{ieeenat_fullname}
    \bibliography{main}
}


\end{document}

%% file: sec/0_abstract.tex
\begin{abstract}
Portrait video editing focuses on modifying specific attributes of portrait videos, guided by audio or video streams. Previous methods typically either concentrate on lip-region reenactment or require training specialized models to extract keypoints for motion transfer to a new identity. In this paper, we introduce a training-free universal portrait video editing framework that provides a versatile and adaptable editing strategy. This framework supports portrait appearance editing conditioned on the changed first reference frame, as well as lip editing conditioned on varied speech, or a combination of both. It is based on a Unified Animation Control (UAC) mechanism with source inversion latents to edit the entire portrait, including visual-driven shape control, audio-driven speaking control, and inter-frame temporal control. Furthermore, our method can be adapted to different scenarios by adjusting the initial reference frame, enabling detailed editing of portrait videos with specific head rotations and facial expressions. This comprehensive approach ensures a holistic and flexible solution for portrait video editing. The experimental results show that our model can achieve more accurate and synchronized lip movements for the lip editing task, as well as more flexible motion transfer for the appearance editing task. Demo is available at \href{https://alice01010101.github.io/RASA/
}{RASA DEMO}.

\end{abstract}

%% file: sec/1_intro.tex
\section{Introduction}
\label{sec:intro}

\begin{figure}[!htbp]
    \centering
    \includegraphics[width=1.0\linewidth]{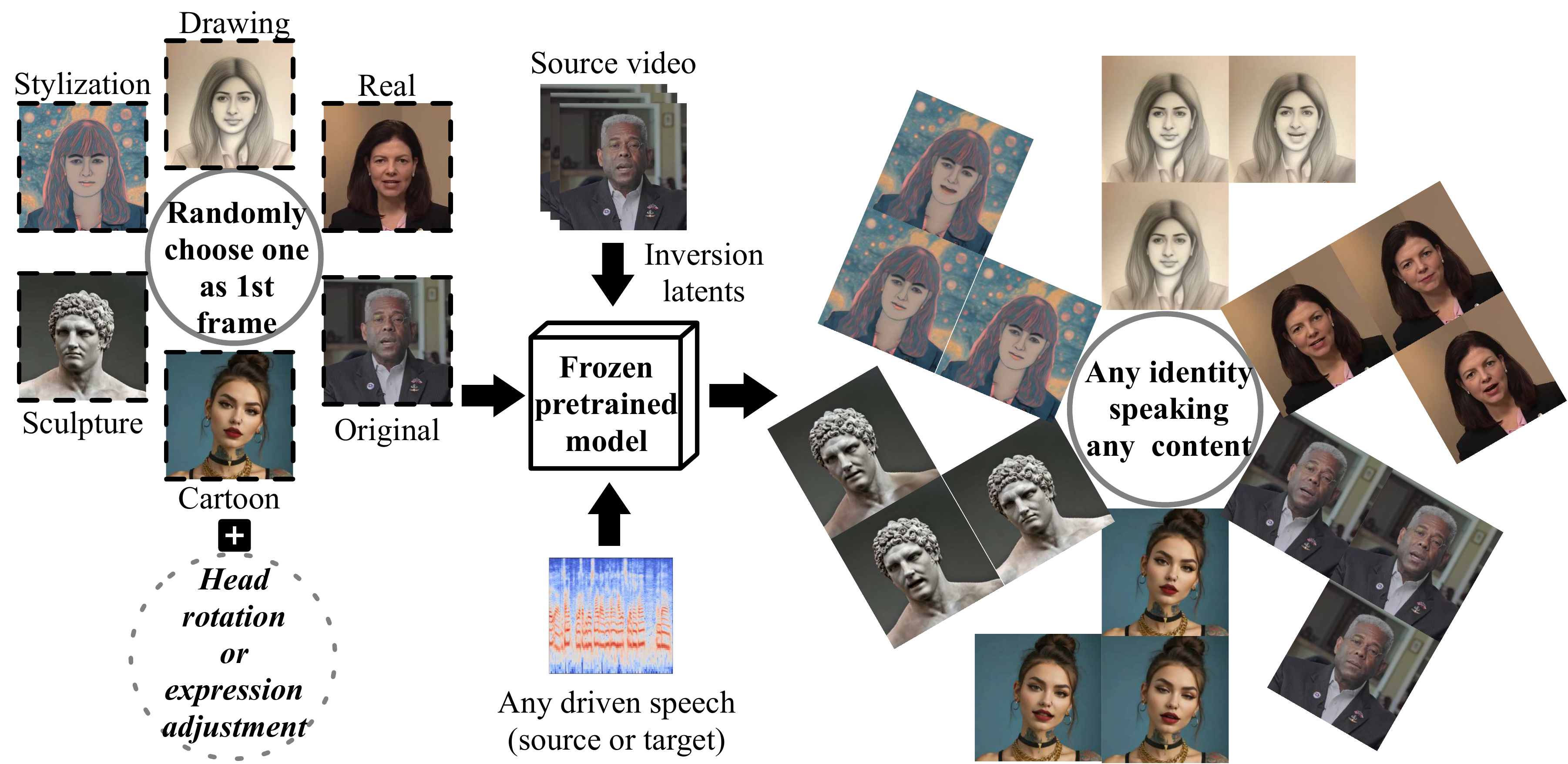}
    \caption{\textbf{The overall pipeline of our model.} RASA enables training-free portrait video editing by leveraging inversion latents of the source video, guided by any portrait identity or speech.}
    \label{fig:anyedit}
\end{figure}

Portrait animation~\cite{sadtalker_2023, emo_2024, hallo2_2024, loopy_2024, aniportrait_2024, xportrait_2024, megactor_2024, vexpress_2024, echomimic_2024, liveportrait_2024, consisid_2024, followyouremo_2024, facechain_2024, lantentsync_2024, ominihuman_2025} aims to generate lifelike videos using a single source image as the appearance reference, optionally deriving motion from driving videos, audio, or text inputs. While most methods focus on enhancing Image-to-Video (I2V) generation by either improving naturalness~\cite{sadtalker_2023, emo_2024, hallo2_2024, loopy_2024, vexpress_2024, followyouremo_2024, emoportrait_2024} or achieving precise motion control~\cite{aniportrait_2024, xportrait_2024, liveportrait_2024, consisid_2024, megactor_2024, megactorsigma_2024}, the area of portrait video editing has received comparatively little research attention. There are instances when it is essential to modify specific aspects of a video while preserving other elements unchanged. Portrait video editing is essential in various fields, including e-commerce product presentations~\cite{Xu_2021, e-commerse_2024} and gaming~\cite{gameai_2024}. It can facilitate rapid adjustments to lip-synchronization for modified speech content or identity transformations, thereby obviating the need for video re-shoots. As shown in \Cref{fig:anyedit}, in this paper, we propose the Replace Anyone, Say Anything (RASA) framework, which is a training-free solution for universal portrait video editing. The framework aims to facilitate changes in identity or speech content, either individually or in combination, to address various needs.

In portrait video editing, both audio and video serve as temporal streams that provide multi-frame information, encompassing both intra-frame and inter-frame relationships. As summarized in \Cref{tab:introduction_table}, we systematically classify portrait video editing methods\footnote{The portrait video editing here refers to portrait animation methods driven by either audio or video streams.} based on their zero-shot adaptation capabilities to changes in the driving modality, including audio, visual, or a combination of both. As for cross-modal audio control, both lip-region movements and overall facial expressions are closely related to specific audio attributes in the driven speech, such as intensity, timbre, and frequency. Existing methods~\cite{videoretalking_2022, musetalk_2024, diffusionvideoediting_2023, lantentsync_2024} focus primarily on lip-region reenactment, which can result in poor audio-visual synchronization and visual information leakage, leading to random lip movements in the absence of audio information. As for visual-driven appearance control, models like LivePortrait~\cite{liveportrait_2024}, MegActor~\cite{megactor_2024}, and X-Portrait~\cite{xportrait_2024} extract keypoints or landmarks to enable motion transfer and maintain coherence with the new portrait identity. MegActor-$\Sigma$~\cite{megactorsigma_2024} extends MegActor to multi-driven generation. These methods typically require training in specialized modules using masking and stitching techniques to enhance head movement coherence between two different identities. However, to tackle challenges related to identity transfer coherence, such as discrepancies in head shape and eye size, these methods are sensitive to variations in portrait video. Additionally, they often prioritize facial keypoint transfer while overlooking background alteration and consistency. \Cref{tab:introduction_table} illustrates that our model excels in supporting various flexible cross-modal tasks. It effectively bridges the gap between different editing objectives while improving overall motion and foreground-background consistency, rather than focusing solely on lip-region reenactment or training a complex network for identity transfer. In particular, our model is based on DDIM inversion to provide additional attention injection, introducing perturbations to alter the original denoising forward trajectory. 

\begin{table}[!t]
    \centering
    \resizebox{1.0\linewidth}{!}{
    \begin{tabular}{c@{\hskip 3pt} | c@{\hskip 3pt} c@{\hskip 3pt} c@{\hskip 3pt} | c@{\hskip 3pt} c@{\hskip 1pt} c@{\hskip 3pt} c@{\hskip 3pt} c@{\hskip 3pt}}
    \toprule
    \multirow{2}{*}{\textbf{Method}} & \multicolumn{3}{c|}{\textbf{Driven}} & \textbf{Task} & \textbf{Edit} & \textbf{Training} & \textbf{Rotation} & \textbf{Expression} \\
    &  A & V & AV & \textit{L} or \textit{S} & \textbf{type} & \textbf{-free} & \textbf{robustness} & \textbf{Adjustment} \\
    \midrule
    
    VideoReTalking~\cite{videoretalking_2022} & \multirow{3}{*}{$\checkmark$} & \multirow{3}{*}{$\times$} & \multirow{3}{*}{$\times$} & \textit{L} & lip & $\times$ & low & \includegraphics[height=1em]{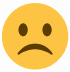} \\
    MuseTalk~\cite{musetalk_2024} &&&& \textit{L} & lip & $\times$ & low & \includegraphics[height=1em]{./pictures/frownie.png} \\
    DVE~\cite{diffusionvideoediting_2023} &&& & \textit{L} & lip & $\times$ & low & \includegraphics[height=1em]{./pictures/frownie.png} \\
    \midrule
    
    LivePortarit~\cite{liveportrait_2024} & \multirow{3}{*}{$\times$} & \multirow{3}{*}{$\checkmark$} &  \multirow{3}{*}{$\times$} &  \textit{S} & keypoints & $\times$ & low & \includegraphics[height=1em]{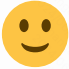} \\
    MegActor~\cite{megactor_2024} &&& & \textit{S} & key areas & $\times$ & low & \includegraphics[height=1em]{./pictures/frownie.png} \\
    X-Portrait~\cite{xportrait_2024} &&& & \textit{S} & keypoints & $\times$ & low & \includegraphics[height=1em]{./pictures/frownie.png} \\
    \midrule
    MegActor-$\Sigma$~\cite{megactorsigma_2024} & $\checkmark$ & $\checkmark$ & $\checkmark$ & both & key areas & $\times$ & low & \includegraphics[height=1em]{./pictures/smiley.png} \\
    \textbf{RASA (Ours)} & \textbf{$\checkmark$} & \textbf{$\checkmark$} & \textbf{$\checkmark$} & \textbf{both} & \textbf{whole face} & \textbf{$\checkmark$} & \textbf{high} & \includegraphics[height=1em]{./pictures/smiley.png} \\
    
    \bottomrule
    \end{tabular}}
    \caption{\textbf{Summary of portrait video editing methods.} In the first row, A and V denote driven audio and video streams respectively, while \textit{L} and \textit{S} represent the lip editing and appearance editing tasks, respectively. By editing the entire face, our method achieves natural and synchronized lip movements, with enhanced robustness and flexibility in handling head rotations and expression adjustments.}
    \label{tab:introduction_table}
\end{table}

The key model component is the Unified Animation Control (UAC) with inversion-based injections. The UAC comprises three components: visual-based shape control, audio-based speaking control, and inter-frame temporal control. First, shape control provides the shape query from the source video for each target frame, serving as implicit motion control for either maintaining motion in the lip editing task or transferring motion in the portrait appearance editing task. Second, the speaking control manages the audio-visual mapping for speech-related lip movements and facial expressions. Lastly, the temporal control enhances inter-frame consistency, balancing video-driven movements and audio-driven movements. Overall, the extensive experimental results show that our method achieves more natural lip movements and better audio-visual synchronization for lip editing than existing methods, driven by changed speech with the same or different identities. It also enables flexible motion transfer for appearance editing and allows for head rotation and expression adjustment by simply editing the initial reference frame.

To summarize, our main contributions are:

\begin{itemize}
    \item We propose the Replace Anyone, Say Anything (RASA) framework, a training-free solution for universal portrait video editing, facilitating independent or joint editing of lip movements and portrait appearance, optionally conditioned on various factors.
    \item We propose the Unified Animation Control (UAC) based on inversion latents, which facilitates shape, speaking, and temporal control for coherent animations, achieving synchronized lip movements and flexible motion transfer.
    \item Our model shows superior experimental results in lip editing, achieving better generation quality and synchronization. It also achieves natural motion transfer results for appearance editing, easily extending to generalized situations with head rotation and expression adjustment.
\end{itemize}

%% file: sec/2_relatedwork.tex
\section{Related Work}
\label{sec:relatedwork}

\subsection{Portrait animation}
Portrait animation encompasses the generation of dynamic facial videos. Some methods~\cite{sadtalker_2023, emo_2024, hallo2_2024, loopy_2024, vexpress_2024, followyouremo_2024, facechain_2024, emoportrait_2024, synctalk_2024} emphasize inter-frame smoothness and natural expression. These techniques leverage neural networks and temporal modeling for fluid transitions. Hallo2~\cite{hallo2_2024} extends Hallo~\cite{hallo_2024} to generate long-duration, high-resolution portrait videos with consistent visual quality and temporal coherence. Conversely, other methods~\cite{aniportrait_2024, xportrait_2024, liveportrait_2024, consisid_2024, megactorsigma_2024, diff2lip_2023, sayanything_2025, joygen_2025} focus on explicit or implicit motion extraction, enabling diverse motion control for practical applications. This allows for precise manipulation of facial dynamics in various contexts, such as animation and virtual reality. AniPortrait~\cite{aniportrait_2024}, which uses explicit keypoints as intermediate motion representations. X-Portrait~\cite{xportrait_2024} directly animates portraits from the original driving video using implicit keypoints for cross-identity training. MegActor~\cite{megactor_2024} animates source portraits with the original video while employing face-swapping and stylization to enhance stability.

\subsection{Inversion-based video editing}
Inversion-based video editing leverages hidden features obtained from the reconstruction branch to enhance the generation process of the target branch. Video-P2P~\cite{videop2p_2023} and Vid2Vid-Zero~\cite{vid2vid-zero_2024} extend the P2P~\cite{p2p_2022} framework into video editing. Video-P2P substitutes spatial self-attention layers with sparse-causal attention, whereas Vid2Vid-Zero emphasizes the need for global spatio-temporal attention layers. FateZero~\cite{fatezero_2023} generates fusion masks for self-attention injection by binarizing cross-attention maps~\cite{blended_2023}. Edit-A-Video~\cite{edit-a-video_2023} interpolates fusion masks between the initial frame and the current frame, using self-attention maps as weights. Make-A-Protagonist~\cite{make-a-protagonistgeneric_2024} follows the approaches of PnP~\cite{pnp_2022} and MasaCtrl~\cite{masactrl_2023}, employing Grounded SAM~\cite{groundedsam_2024} to acquire the segmentation mask for the first frame. UniEdit~\cite{uniedit_2024} and AnyV2V~\cite{anyv2v_2024} separate appearance and motion injection through two auxiliary branches: one for reconstruction and another for motion reference. As far as we know, we are the first to implement inversion-based video editing for animated portrait videos, successfully enabling the simultaneous and flexible transfer of various features.

%% file: sec/3_method.tex
\section{Method}


As illustrated in \Cref{fig:model_structure}, we propose a training-free portrait video editing framework capable of animating any individual to say anything. \Cref{fig:model_structure}(a) outlines the multi-task portrait video editing pipeline. Based on the inversion latents of the source video, we can optionally input the original first frame for lip editing, or replace it with another identity for appearance editing. \Cref{fig:model_structure}(b) details the training-free multi-task portrait editing structure with our proposed Unified Animation Control (UAC). This includes shape control for appearance replacement or maintenance, speaking control for enhancing speech-related visual information, and temporal control to ensure coherence and consistency. In this section, we will first introduce DDIM inversion and the multiple conditional inputs in \Cref{sec:injection}. Next, we will detail the strategy for training-free multi-task portrait video editing in \Cref{sec:UAC}, followed by the processing of the first frame in \Cref{sec:first_frame_processing}.

\subsection{DDIM Inversion and perturbation conditions}
\label{sec:injection}
Diffusion models~\cite{diffusion2_2020, diffusion_2022, diffusion3_2022} generate the desired data samples $z_0$ by iteratively removing noise from initial Gaussian noise $z_{T}$ over $T$ steps. The sampling processes primarily involve Denoising Diffusion Implicit Models (DDIM)~\cite{diffusion_2022, diffusion3_2022} and Denoising Diffusion Probabilistic Models (DDPM)~\cite{diffusion2_2020}. DDIM is more efficient, allowing for step-skipping, while DDPM follows a Markovian process. Our approach combines DDIM sampling with Latent Diffusion Models (LDMs)~\cite{diffusion_2022}, which enhances computational efficiency by operating in a compressed latent space. To generate images from a given $z_{T}$, we use deterministic DDIM sampling~\cite{diffusion3_2022} to iteratively transition from $z_{T}$ to $z_0$.

%

DDIM Inversion, the reverse process of DDIM sampling, is commonly used for editing tasks across various modalities, including images~\cite{inversion1_2022, inversion2_2024}, videos~\cite{make-a-protagonistgeneric_2024, groundedsam_2024, uniedit_2024, anyv2v_2024}, and audio~\cite{audio_editing_2024, music_editing_2024}. While diffusion models excel in feature space~\cite{balaji_2023, dong_2023} for various downstream tasks, applying them to images poses challenges due to the lack of a natural diffusion feature space for non-generated images. Thus, inverting $z_0$ back to $z_{T}$ and saving the latents for the denoising process is essential. DDIM inversion is often employed based on the presumption that the ODE (Ordinary Differential Equation) process~\cite{ODE_2019} can be reversed with sufficiently small step sizes: 
\begin{equation}
\hat{z}_{t} = \alpha \hat{z}_{t-1} + \beta \epsilon_{\theta}(\hat{z}_{t-1}, t-1),
\end{equation}
where $\alpha = \frac{\sqrt{\alpha_t}}{\sqrt{\alpha_{t-1}}}$, $\beta = \sqrt{\alpha_t} \left( \sqrt{\frac{1}{\alpha_t} - 1} - \sqrt{\frac{1}{\alpha_{t-1}} - 1} \right)$. The presumption cannot be guaranteed, and there is often a perturbation between $z_t$ and $\hat{z}_{t}$. In the task of portrait video editing, we propose to produce specific perturbations guided by targeted audio or visual streams. This guidance constrains the perturbations to maintain desired features consistent with previous latent representations while also modifying the target features based on specific additional conditions. In our portrait video editing task, we utilize two key inputs: \textit{the first reference frame}, which serves as the appearance condition, and \textit{the speech}, which acts as the audio condition. These inputs enable us to edit the source video from two distinct perspectives: appearance and speech-related movements, either separately or in combination. Compared with other task-specific methods, our approach offers multi-task, training-free editing, allowing flexible modifications to replace anyone saying anything.


\subsection{Unified animation control}
\label{sec:UAC}
To achieve the multi-task portrait video editing under various conditions, we propose a Unified Animation Control module UAC, shown in \Cref{fig:model_structure}(b). The UAC consists of three components: Shape Control (SC), Cross-modal Speaking Control (CSC), and Temporal Consistency Control (TCC). SC keeps an appearance balance between the source video and the target edited video. CSC facilitates the editing of speech-related facial features, including content-specific lip movements and frequency-related facial details and expressions. TCC further enhances frame-wise consistency and continuity. We apply shape control, speaking cross-modal control, and temporal-motion enhancement at the initial denoising timesteps $\tau_s$, $\tau_a$, and $\tau_f$, respectively.

\noindent\textbf{Shape control.}
Most existing methods~\cite{emo_2024, hallo2_2024, loopy_2024, emoportrait_2024} for generating talking face videos rely on random Gaussian noise, guided by a reference image and speech, which often results in limited motion variation. In this work, we aim to conduct talking-face video editing tasks that generate novel talking-face videos while ensuring consistency with the structural layout established in the original video. MasaCtrl~\cite{masactrl_2023} observed that the layout of objects is primarily formed through self-attention queries. Inspired by this observation, we replace the target query $Q_t$ in the target video generation pipeline with source query $Q_s$, which is derived from the features of the original video. This process is illustrated as follows:

\begin{equation}
    \begin{aligned}
        f_{SC}(t) := 
        \begin{cases}\{Q_{\textrm{s}},K_{\textrm{t}},V_{\textrm{t}}\}&t \geq \tau_s \\ \{Q_{\textrm{t}},K_{\textrm{t}},V_{\textrm{t}}\} & t < \tau_s\end{cases}\\
    \end{aligned}
    \label{tab:shape_control}
\end{equation}

The $Q_s$, $K_s$, and $V_s$ represent the query, key, and value from the source original video for the SC module. We replace only $Q_{t}$ instead of both $Q_{t}$ and $K_{t}$ because the attention map $M_{s}$ from the original video contains excessive portrait information, which can result in significant identity distortion. We inject the $Q_s$ for the initial $\tau_s$ steps to strike a balance between altering identity and enabling effective feature adaptation. By maintaining structural layout consistency between the source and target talking face videos, we preserve essential facial features and spatial arrangements, thereby enhancing realism and preventing distortions.

\noindent\textbf{Cross-modal speaking control.} Depending on practical needs, we may require different portraits to either speak the same words or different words. The cross-attention mechanism between visual and audio features provides a direct method to establish the mapping between audio and the corresponding facial features. In the context of cross-modal speaking control, we inject the source visual query $Q_{s}^v$ and source audio key $K_{s}^a$ obtained from the DDIM inversion into the target generation pipeline to replace the target visual query $Q_{t}^v$ and target audio key $K_{t}^a$ during the initial $\tau_a $ denoising steps, as denoted below:
\begin{equation}
    \begin{aligned}
        f_{CSC}(t) := 
        \begin{cases}\{Q_{\textrm{s}}^v,K_{\textrm{s}}^a,V_{\textrm{t}}^a\}&t \geq \tau_a \\ \{Q_{\textrm{t}}^v,K_{\textrm{t}}^a,V_{\textrm{t}}^a\} & t < \tau_a\end{cases}\\
    \end{aligned}
    \label{tab:speaking_control}
\end{equation}

This approach ensures a consistent mapping between audio inputs and corresponding lip movements and facial expressions. Such alignment enhances the coherence between auditory and visual components, facilitating more natural interactions and improving the realism of the generated output. In our video editing task, we maintain coherence in lip movements and facial expressions for a portrait video when the audio is the same and the spoken words are identical. Conversely, when the driven audio differs and the spoken words change, we adapt the lip movements and facial expressions to achieve a more natural performance.

\noindent\textbf{Temporal consistency control.}
%
Finally, we enhance the frame-wise motion consistency of the target video. This process improves previously edited sections to align with the new appearance or speech conditions in the SC and CSC. It ensures that these changes remain consistent with the original movements in each frame. The details for controlling temporal consistency are similar to those for speaking control. Specifically, we replace the query $Q_{\textrm{t}}^{f_m}$ for the $m$-th target frame with the query $ Q_{\textrm{s}}^{f_m} $ for the $m$-th source frame. And we replace the key $K_{\textrm{t}}^{f_n}$ for the $n$-th target frame with the key $K_{\textrm{s}}^{f_n}$ for the $n$-th source frame. This allows us to inject the source video's frame-wise attention into the target video.

Overall, the SC, CSC, and TCC modules are capable of supporting a wide range of editing needs, either individually or in combination. SC maintains movement consistency for the same identity and facilitates motion transfer across different identities. With SC, we can adjust the inversion injection timesteps to maintain the shape and motion during the lip editing process, as demonstrated in the ablation results in \Cref{fig:ablation_for_pnps_pnpt}(a). This adjustment also enhances the transfer of shape and implicit movements from the source portrait to the target portrait in the appearance editing task. CSC preserves lip movements and facial details for identical speech content while adapting to corresponding lip movements and facial expressions for altered speech content. This integration ensures a more accurate and nuanced representation of speech and expression in the animated portraits. TCC enhances coherence between audio-driven and visual-driven movements, ensuring overall motion consistency. As illustrated in \Cref{fig:model_structure}(a), through the proposed module integration, we can achieve flexible portrait video editing that enables the animation of any individual speaking any content.
 
\subsection{First frame processing for appearance editing}
\label{sec:first_frame_processing}
In this subsection, we detail the first frame processing module, as illustrated in \Cref{fig:model_structure}(a). The first reference frame is crucial during the denoising process, as it provides both pixel-level information and clip-based embeddings. As shown in \Cref{fig:anyedit}(a), it serves as the appearance condition for both portrait appearance editing and lip editing in our task. Processing the reference frame for diverse styles and identities enables adaptation to a wide range of realistic scenarios. To alter the first reference frame, we utilize several techniques: 1) To modify the portrait while preserving the background, we utilize a face-parsing method~\cite{bisenet_2018} to isolate the foreground portrait from the target image and the background from the source image. These components are then seamlessly merged, with inpainting techniques~\cite{physgen_2024} applied to address any inconsistencies, ensuring a natural integration. 2) To generate portrait videos from diverse viewpoints, we employ the approach described in \cite{facevid2vid_2021} to synthesize new head poses with varying yaw, pitch, and roll angles. 3) For creating portrait videos with nuanced facial expressions, we leverage techniques from \cite{liveportrait_2024} to adjust the openness of the eyes and mouth. It is important to note that our current work does not address finer facial expression details, such as eyebrow movement, eyeball shifts, teeth visibility, or Ajna region modifications, which are left for future investigation.

%% file: sec/4_experiments.tex
\section{Experiments}
\label{sec:experiments}

\begin{table*}
    \centering
    \resizebox{1.0\textwidth}{!}{
        \begin{tabular}{cccccc | cccc}
             \toprule
             \multirow{2}{*}{Method} & \multicolumn{5}{c|}{Self-id Driven Audio} & \multicolumn{4}{c}{Cross-id Driven Audio} \\
             & FID$\downarrow$ & FVD$\downarrow$ & LPIPS$\downarrow$ & LSE-D$\downarrow$ & LSE-C$\uparrow$ & NIQE$\uparrow$ & CSIM$\uparrow$ & LSE-D$\downarrow$ & LSE-C$\uparrow$ \\
             \midrule
             Wav2Lip~\cite{wav2lip_2020} & 35.727 & 225.706 & 0.257 & \textbf{6.384} & \textbf{8.049} & \underline{5.716} & 0.811 & 8.152 & \underline{7.205} \\
             VideoReTalking~\cite{videoretalking_2022} & 30.834 & 208.496 & 0.256 & 8.082 & 6.871 & 5.508 & 0.756 & 8.664 & 6.458 \\
             TalkLip~\cite{talklip_2023} & 40.061 & 264.723 & 0.257 & 10.111 & 4.549 & 5.632 &  0.799 & 10.212 & 4.813 \\
             DI-Net~\cite{dinet_2023} & 34.130 & 231.347 & \underline{0.250} & 9.167 & 7.423 & 5.385 & 0.771 &  \underline{8.147} & 6.923  \\
             DVE~\cite{diffusionvideoediting_2023} & - & - & - & - & - & 5.192 & - & 12.569 & 0.825 \\
             Musetalk~\cite{musetalk_2024} & \underline{24.419} & \underline{197.162} & 0.271 & 10.360 & 3.904 & 5.480 & \textbf{0.839} & 10.560 & 4.509\\
             Ours & \textbf{23.327} & \textbf{188.611} & \textbf{0.248} & \underline{6.975} & \underline{7.875} & \textbf{6.096} & \underline{0.823} & \textbf{7.777} & \textbf{7.689} \\
             \bottomrule
        \end{tabular}
    }
    \caption{\textbf{Portrait lip editing results for audio-based portrait video editing.} The quantitative comparisons are conducted on the HDTF dataset. The self-ID driven audio section involves altering the audio from the same identity but using different audio segments. The cross-ID driven audio section entails changing the audio from cropped segments of different identities. The best results are in \textbf{bold} and the second best results are \underline{underline}.}
    \label{tab:tab_lip_reeactment}
\end{table*}

\begin{figure*}
    \centering
    \includegraphics[width=1.0\linewidth]{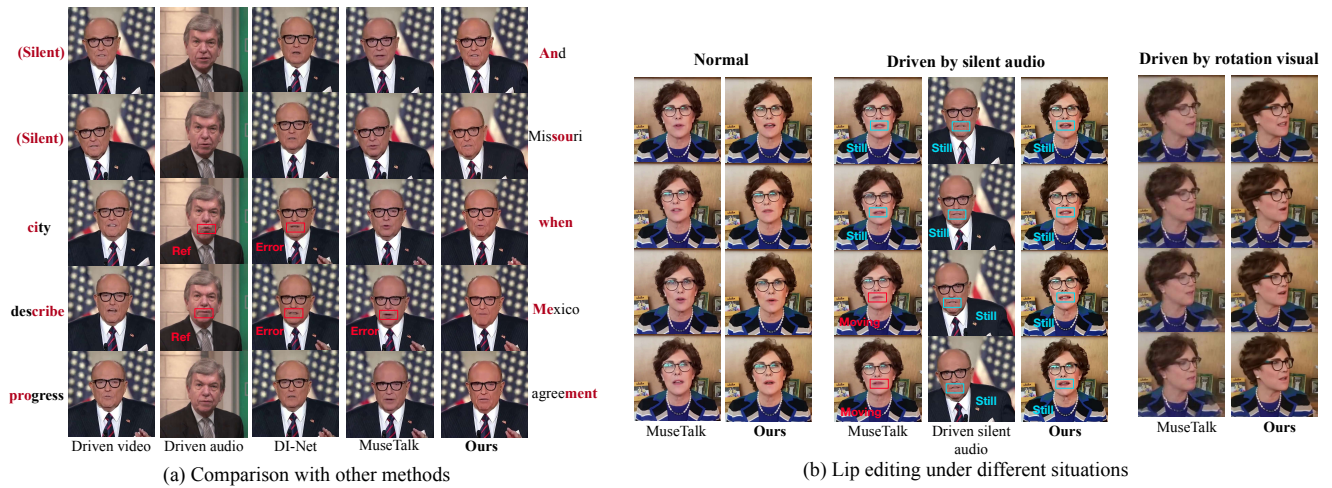}
    \caption{\textbf{Qualitative comparisons for the portrait lip editing.} We selected two edited video samples from the HDTF dataset to compare our methods with others. In part (a), our audio-driven approach shows more natural lip movements and better synchronization. Part (b) highlights the robustness of our results in various scenarios, including silent speech and head rotations.}
    \label{fig:ReTalking_visualization} 
\end{figure*}

\begin{figure*}
    \centering
    \includegraphics[width=1.0\linewidth]{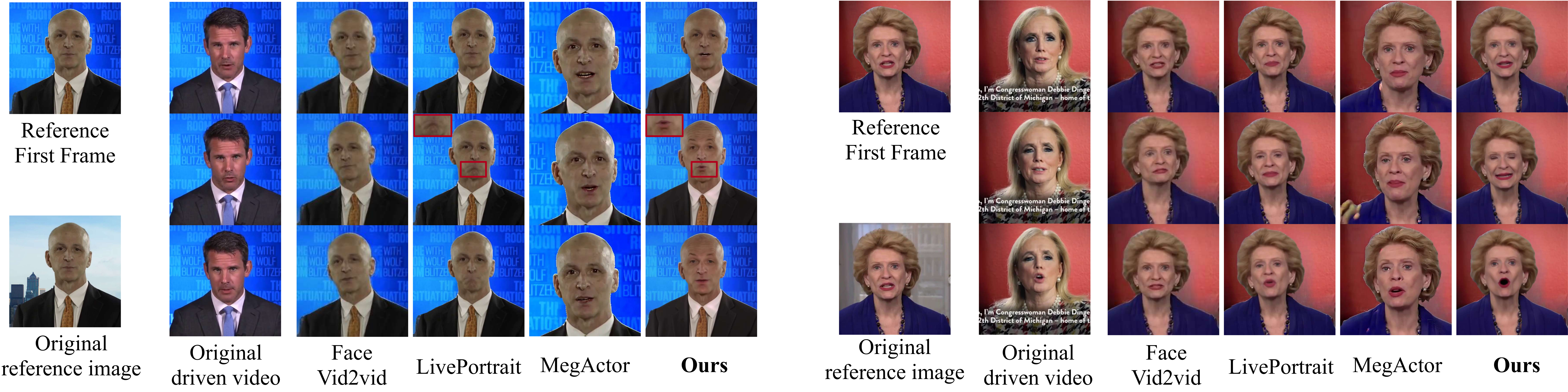}
    \caption{\textbf{Qualitative comparisons for the task of portrait appearance editing.} To achieve portrait video appearance editing, we first apply background inpainting to obtain an edited first frame as the reference frame, using the target portrait as the foreground and the source background. We then input this reference image into our model to implement the portrait video appearance editing.}
    \label{fig:headchange_visualization}
\end{figure*}


\subsection{Experimental setups}
\paragraph{Datasets.} To evaluate our proposed method, we utilized the publicly available dataset HDTF~\cite{HDTF_2021}. The HDTF dataset comprises approximately 410 in-the-wild videos, each available in 720P or 1080P resolution. For data processing, we follow the approach outlined in \citet{anitalker_2024} and enlarge the cropping region to include the shoulders. For a fair comparison, followed Wav2Lip~\cite{wav2lip_2020} and VideoReTalking~\cite{videoretalking_2022}, we randomly selected 20 videos, each containing 160 frames (6.4 seconds) in the testing stage. Specifically, during video editing, the reference image is randomly selected from one frame of the current video, while the driven audio is sourced from a different video. We further categorized the process into two scenarios: self-ID driven audio, where the audio originates from the same individual, and cross-ID driven audio, where the audio comes from a different person. This distinction is essential, as individuals have distinct timbres and pitches, introducing additional complexity to the evaluation process.

\noindent \textbf{Evaluation metrics.} To evaluate visual fidelity, identity preservation, and lip synchronization capabilities, we employ multiple evaluation metrics, which are introduced as follows:
We assess our model using both image- and video-based generative metrics, including FID~\cite{fid_2018}, FVD~\cite{fvd_2019}, and LPIPS~\cite{lpips_2018}. These metrics enable us to evaluate the generative quality of audio-driven face-changing videos in comparison to ground-truth videos from various perspectives. LSE-D and LSE-C~\cite{wav2lip_2020} specifically measure lip synchronization through lip-sync error distance and confidence levels. For scenarios involving cross-ID driven audio, where the audio is sourced from a different individual, we utilize NIQE~\cite{niqe_2012} to assess the generated image quality without reference images. Additionally, we employ CSIM to evaluate cross-ID similarity, focusing on identity preservation in the cross-ID setting.

\noindent \textbf{Implementation details.} We first perform DDIM inversion for 500 backward steps with classifier-free guidance set to 1.0. The forward denoising steps are configured to 50, with a guidance scale of 3.5. The entire inference process is carried out on a single NVIDIA Tesla V100. For the default settings, we configure the injection steps for shape control, denoted as $\tau_{s}$, speaking control, represented as $\tau_{a}$, and temporal control, indicated by $\tau_{f}$, to values of 0.2, 0.2, and 0.4, respectively. The parameters $\tau_{s}$ and $\tau_{a}$ are detailed in \Cref{tab:shape_control} and \Cref{tab:speaking_control}, respectively. To facilitate progressive video editing, we send every 16 frames to the denoising forward process. To mitigate error accumulation during the iterative generation process, we apply a random mask with a strength of 0.25 to the last two generated frames before concatenating them with the next set of 16 frames.

\noindent \textbf{Compared baselines.} The proposed method is compared with several state-of-the-art video dubbing methods: 
1) Wav2Lip~\cite{wav2lip_2020} excels in realistic lip synchronization using a pre-trained discriminator.
2) VideoReTalking~\cite{videoretalking_2022} offers high-quality audio-driven lip-syncing with expression neutralization and speaker identity awareness.
3) DI-Net~\cite{dinet_2023} produces photorealistic videos by integrating a dual-encoder framework focused on facial action units.
4) TalkLip~\cite{talklip_2023} employs contrastive learning and a transformer model for improved temporal alignment between audio and video.
5) MuseTalk~\cite{musetalk_2024} generates high-fidelity lip-sync targets in a latent space via a Variational Autoencoder for efficient video creation.

\subsection{Results on portrait lip editing}
The timbre, pitch, and loudness of audio are closely linked to lip movements and facial expressions. Most existing lip editing methods, such as VideoReTalking~\cite{videoretalking_2022} and MuseTalk~\cite{musetalk_2024}, primarily focus on masking and regenerating the mouth region, often overlooking necessary adjustments to ensure the consistency and naturalness of the entire face. As shown in \Cref{tab:tab_lip_reeactment}, we evaluate the lip editing performance on 20 randomly selected samples from the HDTF dataset. This evaluation involves replacing the original audio with both self-identity-driven and cross-identity-driven audio. For self-identity lip editing, we compare the generated videos against the same-audio-driven ground truth using metrics such as FID and FVD to evaluate the quality of individual frames and frame-by-frame videos, along with LPIPS to measure perceptual similarity. We also assess audio-visual synchronization using LSE-D distance and LSE-C confidence. For cross-identity lip editing, since ground-truth generated videos are not available, we rely solely on the no-reference metric NIQE for evaluation. The results indicate that our method outperforms existing approaches across multiple metrics, achieving a more accurate and natural representation while effectively balancing motion consistency and naturalness.

The qualitative visualization results are presented in \Cref{fig:ReTalking_visualization}. As shown in \Cref{fig:ReTalking_visualization}(a), our method effectively preserves the shape and implicit pose motion when compared to the generation results from DI-Net~\cite{dinet_2023}. Furthermore, in comparison to the generation results from MuseTalk~\cite{musetalk_2024}, our approach demonstrates superior audio-lip synchronization and improved lip-face consistency. Specifically, our model employs shape control and temporal control to ensure motion consistency for intra-frame shapes and frame-wise videos, respectively. Additionally, speaking control leverages cross-modal audio mapping to adjust the corresponding lip movements, enhancing lip-face coherence and improving the overall naturalness of the generated output. In \Cref{fig:ReTalking_visualization}(b), we illustrate the superiority of our model under various conditions, including silent speech and head rotation. When the provided speech is silent, MuseTalk tends to produce random lip movements due to visual information leakage, whereas our model maintains stillness in the lips. Furthermore, our approach achieves greater coherence and improved lip movements during head rotation situations. 

\subsection{Results on portrait appearance editing}
As shown in \Cref{fig:headchange_visualization}, we present qualitative results for this process. Unlike LivePortrait~\cite{liveportrait_2024} and MegActor~\cite{megactor_2024}, which focus on extracting explicit or implicit keypoints for precise face motion transfer, our method utilizes DDIM inversion noise latents to inject implicit motion features into the current denoising pipeline.
From \Cref{fig:headchange_visualization}, it is evident that mapping intermediate keypoints or landmarks to another identity does not always succeed. Our qualitative results demonstrate that our method extends the consistency of face keypoints to encompass broader modifications, including the head, hair, and clothing, resulting in more natural and versatile portrait transformations. 
%
%
Compared to the video editing results from LivePortrait~\cite{liveportrait_2024}, our approach achieves more accurate and detailed generation, as LivePortrait exhibits some blurring in the lip region indicated by the red box in \Cref{fig:headchange_visualization}.
Additionally, when compared to MegActor~\cite{megactor_2024}, we maintain a closer identity resemblance to the original reference image.

\begin{figure}[!htbp]
    \centering
    \includegraphics[width=0.90\linewidth]{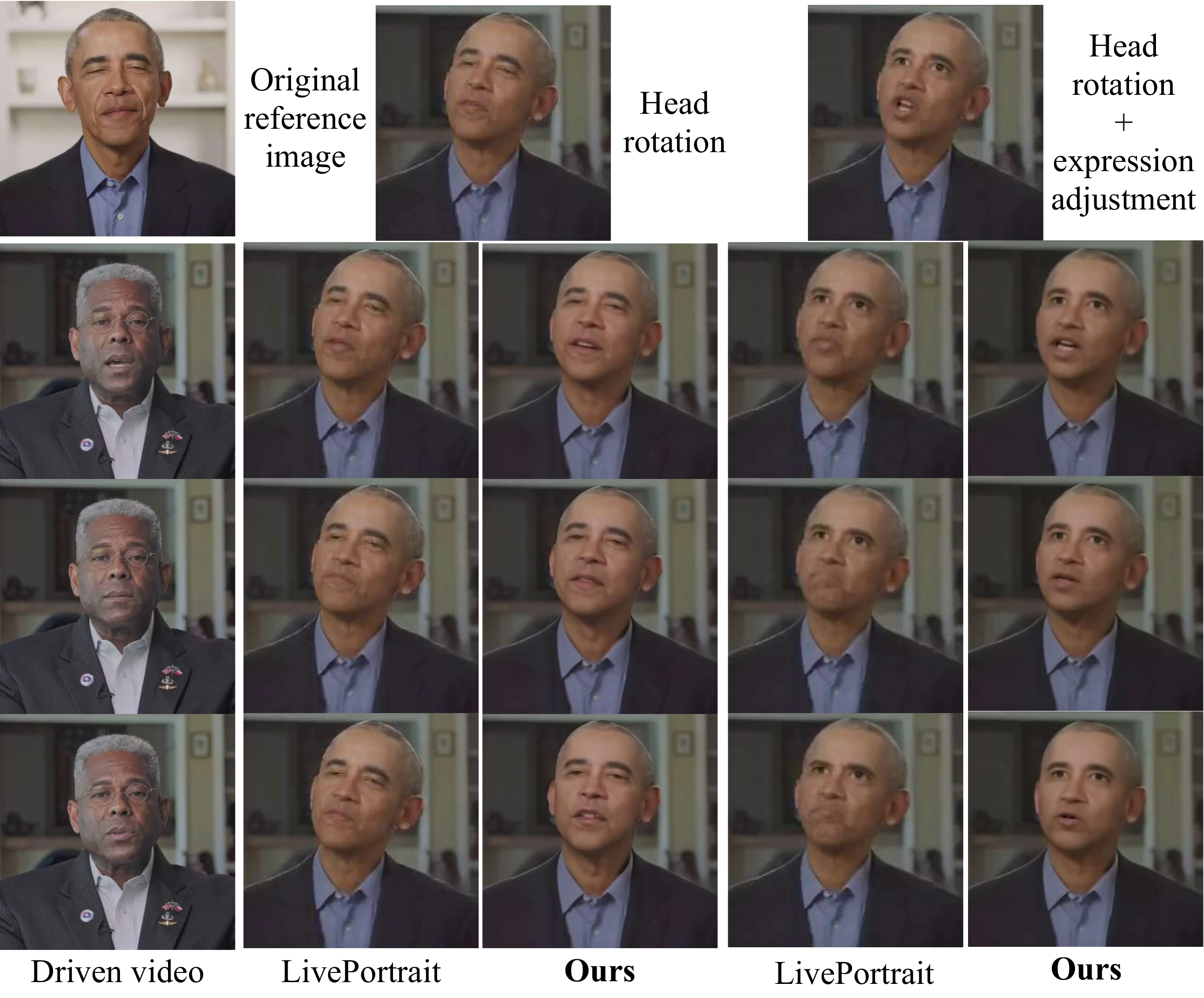}
    \caption{\textbf{Portrait appearance editing with different views.} From left to right, we sequentially edit the first reference image by changing the background, applying head rotations, and expression editing. }
    \label{fig:free_views}
\end{figure}

We further demonstrate the robustness and flexibility of our model for appearance editing under head rotation and expression editing. As shown in \Cref{fig:free_views}, we demonstrate the edited results with head rotation and expression editing. Compared to LivePortrait, which strictly maintains rotation angles and keypoint alignment, our method produces more natural and coherent edited videos. Overall, unlike other portrait appearance editing methods~\cite{liveportrait_2024, megactor_2024, xportrait_2024, megactorsigma_2024} that emphasize rigid alignment, our approach prioritizes fluidity and realism, enabling greater artistic expression and adaptability across various scenarios. More qualitative results can be found in the supplementary materials.

\subsection{Ablation study}
\begin{figure}
    \centering
    \includegraphics[width=1.0\linewidth]{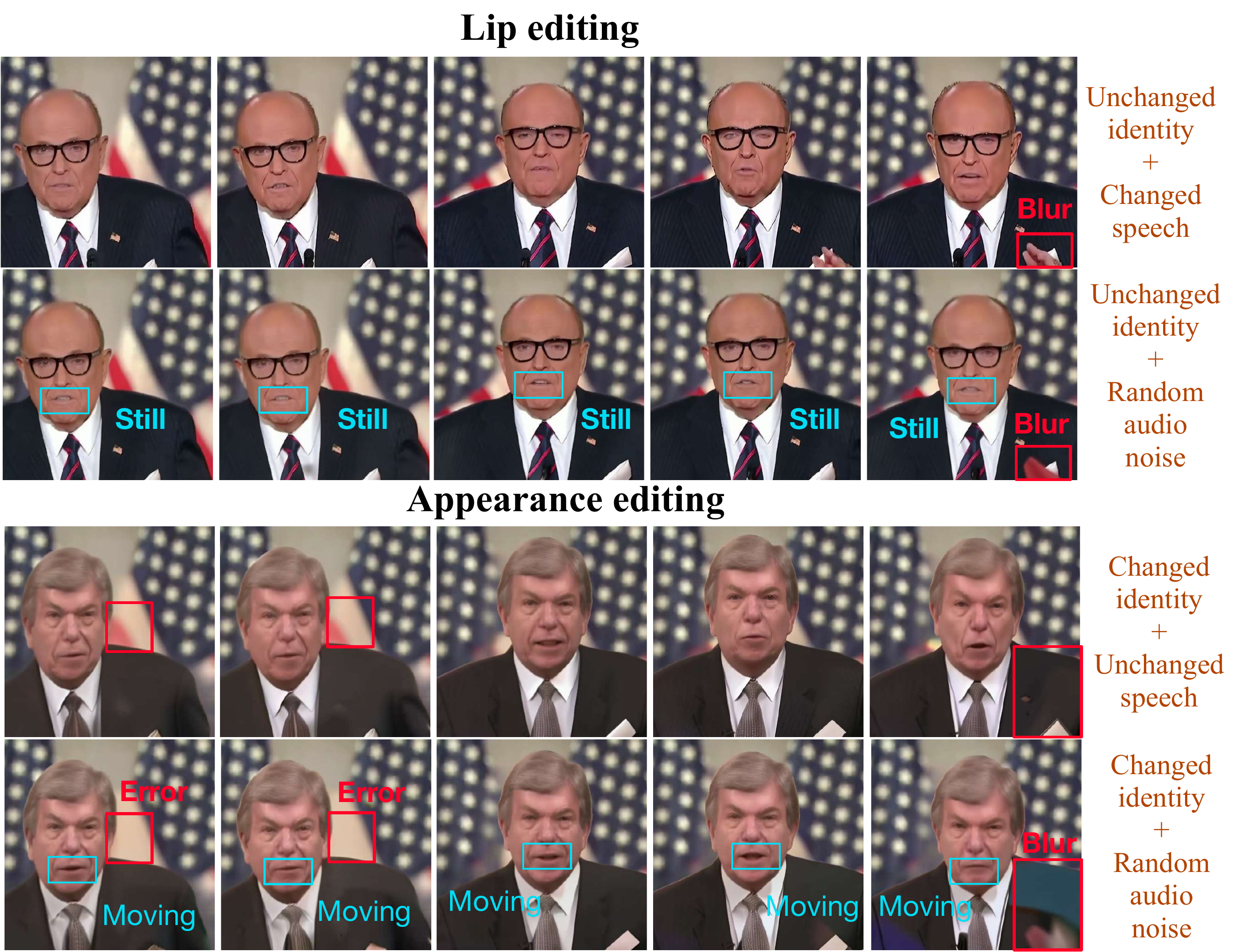}
    \caption{\textbf{Ablation study for the driven audio.} The red box emphasizes that speech maintains image details like backgrounds and facial features. The cyan box shows that while our model effectively maps speech to lip movement, there is still potential for improvement in handling identity and speech changes.}
    \label{fig:ablation_for_multi_task}
\end{figure}

\begin{figure}
    \centering
    \includegraphics[width=1.0\linewidth]{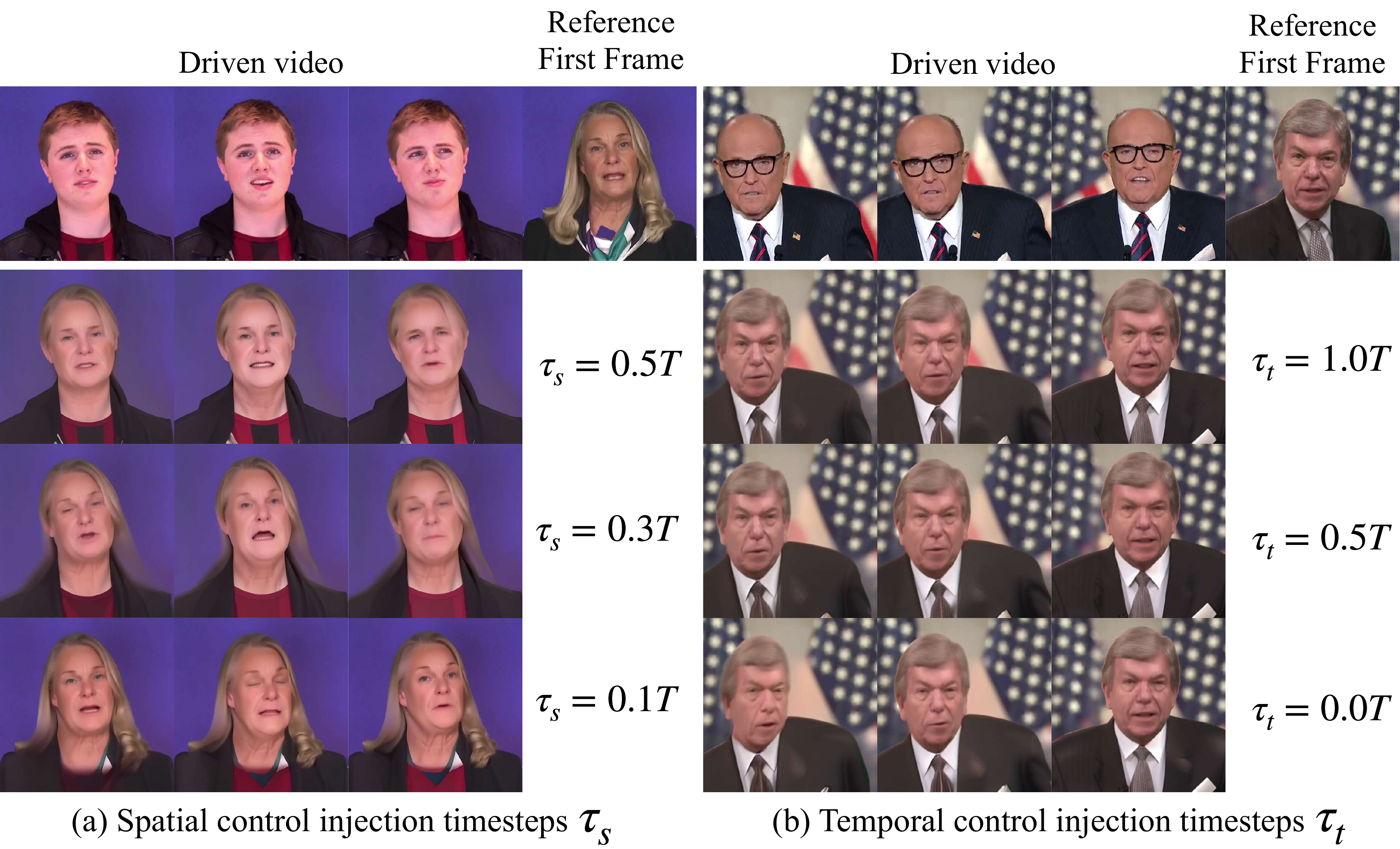}
    \caption{\textbf{Ablation study for spatial control injection timesteps $\tau_s$ and temporal control injection timesteps $\tau_t$.} We set the initial injection timesteps as follows: for spatial control $\tau_s$, we used 0.5, 0.3, and 0.1 of total denoising timesteps $T$; for temporal control $\tau_t$, we applied 1.0$T$, 0.5$T$, and 0.0$T$.}
    \label{fig:ablation_for_pnps_pnpt}
\end{figure}


\paragraph{The effectiveness of driven audio.} In \Cref{fig:ablation_for_multi_task}, we investigate the effectiveness of driven audio from two perspectives. First, in both the lip editing and appearance editing tasks, we observe that incorporating speech significantly enhances the preservation of image details, such as the background, hands, and shoulders. In contrast, using random audio noise results in blurring, as highlighted by the red boxes. Second, in the lip editing task (indicated by the cyan box), we find a strong correlation between speech and lip movements; when random audio noise is used, the mouth often remains open. However, during the video appearance editing task, our model struggles to manage changes in both identity and speech simultaneously. This highlights the need for further research into how random audio noise affects lip movements. 

\noindent\textbf{Shape and temporal control injection.} As shown in \Cref{fig:ablation_for_pnps_pnpt}(a), we set the shape control injection steps, denoted as $\tau_{s}$, to 0.5, 0.3, and 0.1 of the total denoising timesteps $T$, respectively. The visualization results indicate that a higher degree of shape query injection leads to the generated target portrait shape being more similar to the source portrait. This method maintains a balance between pose and facial expression consistency while preserving the fidelity of the provided reference portrait image. Regarding the temporal control injection, illustrated in \Cref{fig:ablation_for_pnps_pnpt}(b), we configured the initial injection inversion timesteps to $0.0T$, $0.5T$, and $1.0T$. When $\tau_t$=$0.0T$, it indicates the absence of the temporal attention injection module, resulting in noticeable blurring or distortion during large-scale movements in the source-driven video. Additionally, as the temporal injection timesteps increase from $0.0T$ to $1.0T$, we observe that more background details are preserved, including the intricate features of the flag in this example. We set $\tau_s$ to 0.2$T$ and $\tau_t$ to 0.4$T$ in our experiment.  

\begin{table}[!h]
    \centering
    \begin{tabular}{cccc}
        \toprule
        $\tau_a$ & FVD $\downarrow$ & LSE-D $\downarrow$  & LSE-C $\uparrow$ \\
        \midrule
        0.2T & 190.504 & 7.406 & 7.541 \\
        0.5T & 190.097 & \textbf{7.112} & \textbf{7.766} \\
        0.8T & \textbf{189.213} & 7.359 & 7.698 \\  
        \bottomrule
    \end{tabular}
    \caption{\textbf{Ablation study for the speaking control injection timesteps $\tau_a$.} For the audio-visual cross-modal attention map injection, we evaluate our model using the FVD metric for inter-frame details fidelity, as well as the LSE-D and LSE-C metrics for audio-visual synchronization. }
    \label{tab:ablation_for_pnpcross}
\end{table}
\vspace{-1pt}

\noindent\textbf{Speaking control injection.} The speaking attention maps demonstrate the relationship between video and speech across modalities. In a Self-id driven audio setting, we replaced the driven speech with segments from the same speaker. The results, as shown in \Cref{tab:ablation_for_pnpcross}, indicate that increasing the speaking injection timesteps from $0.2T$ to $0.8T$ improves the FVD metric. This improvement suggests that the speaking attention map effectively captures detailed visual information related to speech, including both facial features and background details. Additionally, audio-visual synchronization, measured by LSE-D and LSE-C metrics, improves as $\tau_a$ increases from $0.2T$ to $0.5T$, suggesting enhanced audio-visual mapping. However, synchronization declines when $\tau_a$ increases from $0.5T$ to $0.8T$, likely due to excessive injection of original speech-related lip information, which may disrupt the editing process.

%% file: sec/5_conclusion.tex
\section{Conclusions}

In this paper, we propose a training-free framework for universal portrait video editing that allows anyone to say any content. Our method facilitates flexible lip and appearance editing through UAC, driven by altered speech or visual identity. UAC provides control over shape, speech, and temporal based on inversion latents. Additionally, it supports background changes, head rotation, and expression adjustments, ensuring adaptability to various scenarios. In the future, we will explore multi-language editing scenarios and investigate motion transfer between identities with significant differences, such as variations in head size under close-up and long-shot conditions.  